\documentclass[11pt,a4paper]{article}

\usepackage[margin=1in]{geometry}
\usepackage{hyperref}
\usepackage{booktabs}
\usepackage{graphicx}
\usepackage{array}
\usepackage{amsmath}
\usepackage{amssymb}
\usepackage{url}
\usepackage{float}
\usepackage{algorithm}
\usepackage{algpseudocode}
\usepackage[protrusion=true,expansion=false]{microtype}

\hypersetup{colorlinks=true,linkcolor=black,urlcolor=blue,citecolor=blue}

\PassOptionsToPackage{hyphens}{url}
\makeatletter
\g@addto@macro\UrlBreaks{\UrlOrds}
\g@addto@macro\UrlBreaks{\do\_\do\/\do\-\do\.\do\:\do\(\do\)\do\[\do\]\do\,}
\makeatother

\title{ConvMemory v2: A Recall-Preserving Top-10 Evidence Reranker \\
       for Conversational Memory Retrieval}

\author{Taiheng Pan \\
        School of Computing and Information Systems \\
        University of Melbourne \\
        \texttt{github.com/pth2002}}

\begin{document}

\maketitle

\begin{abstract}
We describe ConvMemory v2, an opt-in token-evidence reranker that sits after the
lightweight ConvMemory v1 reranker and reorders only v1's protected top-10
candidate set. v2 is a fine-tuned \texttt{ms-marco-MiniLM-L-6-v2} cross-encoder
(22{,}713{,}601 parameters, measured from the released checkpoint) applied to the
ten (query, memory) pairs that v1 has already selected; it does not change which
ten memories are returned, so Recall@10 and Hit@10 are identical to v1 by
construction, not by statistical coincidence. On the LoCoMo conversational
memory benchmark (5 seeds, $n=4955$ test rows), v2 raises FULL MRR from v1's
0.5824 to 0.6560 (paired bootstrap $+0.0734$, 95\% CI $[+0.0645, +0.0827]$) and
H@1 from 0.4440 to 0.5474. v2 closes most but not all of the gap to a much more
expensive full-pool cross-encoder reference (mxbai-rerank-large-v1 over the
top-500, MRR 0.6688): on FULL MRR v2 sits 0.013 below mxbai\_top500, but on two
raw-dense-hard slices (where v1's protected top-10 has higher recall than mxbai's
own top-10) v2 exceeds mxbai\_top500. A four-arm load-bearing ablation shows
candidate-specific memory text is the mechanism: removing, shuffling, or
replacing it collapses MRR below raw dense retrieval. v2 is best understood as a standard recall-preserving cascade pattern with
LoCoMo-specific fine-tuning, an explicit anti-shortcut inference contract, and
disciplined load-bearing analysis; its advantage over mxbai is slice-specific
rather than a general dominance claim. This report extends
the v1 technical report \cite{convmemory_v1}.
\end{abstract}

\newpage
\setcounter{tocdepth}{2}
\tableofcontents
\newpage

\section{Introduction}
\label{sec:intro}

The v1 ConvMemory technical report \cite{convmemory_v1}
studied how cheaply a small learned reranker could approximate cross-encoder
quality on conversational long-term memory retrieval. Its default reranker
deliberately avoids running a per-query, per-candidate transformer forward over
the candidate pool (v1 \S3.4); it organizes a high-recall top-500 pool cheaply.
A natural follow-up question is: once v1 has already narrowed the pool to a small,
high-recall prefix, can a small, bounded amount of token-level cross-encoder
computation, spent only on that prefix, improve ordering without sacrificing
recall and without becoming as expensive as a full-pool cross-encoder?

This report answers that question with ConvMemory v2, a recall-preserving
top-10 evidence reranker. v2 takes the exact top-10 set that v1 returns, scores
those ten (query, memory) text pairs with a fine-tuned
\texttt{ms-marco-MiniLM-L-6-v2} cross-encoder, reorders the ten candidates by
that score, and appends v1's unchanged tail. Because the top-10 \emph{set} is
preserved, Recall@10 and Hit@10 are unchanged from v1 by construction.

We make three contributions.

\begin{enumerate}
    \item \textbf{ConvMemory v2}: an opt-in,
    recall-preserving top-10 evidence reranker distributed as a released
    checkpoint on the Hugging Face Hub. Its value is a new point on the
    cost--quality frontier: on LoCoMo it recovers a large fraction of the MRR
    and H@1 headroom above v1 at roughly $1.7\times$ v1's latency, while
    remaining about $68\times$ cheaper than a full-pool cross-encoder in our
    measurement. On FULL MRR, v2 remains below mxbai-rerank-large-v1; its advantage over
    the full-pool reference is slice-specific (\S\ref{sec:results}).
    \item \textbf{A mechanism-level ablation}: a
    four-arm, 5-seed token-evidence ablation showing that candidate-specific
    memory text is what drives v2's gain. Removing the memory text, shuffling it
    within a question, or replacing it with text from other questions does not
    merely erase the gain --- it drives MRR below raw dense retrieval, the
    cleanest available fingerprint that token-on-memory-text alignment is
    load-bearing.
    \item \textbf{An explicit anti-shortcut inference contract}: the released
    inference API rejects a fixed set of
    gold-defining and teacher-derived fields, and train/test conversations are
    disjoint. This is backed by machine-checkable tests, not just prose.
\end{enumerate}

v2 builds on a recall-preserving cascade, an established information-retrieval
pattern; its contribution lies in the LoCoMo-specific fine-tuning, the
recall-preserving design, the load-bearing analysis, and the anti-shortcut
inference contract. All results reported here are on LoCoMo at the retrieval
stage.

\section{Relationship to the v1 Paper}
\label{sec:rel}

Because v2 composes with v1 rather than replacing it, and because the v1 report
made a specific structural cost claim and a specific negative attribution
result, we make the relationship explicit before describing v2. This section
expands the ``Relationship to the v1 paper'' note in the public v2
documentation.\footnote{\url{https://github.com/pth2002/ConvMemory/blob/main/docs/EVIDENCE_RERANKER.md}}

\paragraph{v1 \S3.4 (the ``no per-pair transformer forward'' structural claim).}
v1's core cost argument is that its default path does not run a transformer
forward over each query--candidate pair in the pool. v1 alone still honours this:
\texttt{retrieve(query, memories)} uses the pure v1 path unless v2 is explicitly
requested. v2 relaxes this claim only on the protected top-10:
it scores ten query--memory pairs per query (one per protected candidate), not 500. v1 and v1+v2 are therefore two points on the same cost--quality frontier,
complementary points rather than substitutes.

\paragraph{v1 \S3.3 (teacher choice discipline).} v1 deliberately did not use the
strongest available cross-encoder (mxbai-rerank-large-v1) as its distillation
teacher, to avoid conflating distillation gains with teacher choice. v2's
headline arm preserves this discipline: it is trained with a gold-only listwise
objective (teacher weight $0.0$; \S\ref{sec:training}). A cross-encoder-teacher
variant exists but is not the headline and is not load-bearing
(\S\ref{sec:results}).

\paragraph{v1 \S5 (temporal window not load-bearing).} v1 published a negative
result: its learned temporal window is statistically significant on aggregate
but not temporally specific. v2's supported mechanism is distinct from the temporal window and is tested
separately (\S\ref{sec:ablation}): candidate-specific memory text is load-bearing.
The v1 negative result stands unchanged.

\paragraph{v1 \S7 (from-scratch stream rerankers fail without a teacher signal).}
v1 \S7 showed that small from-scratch stream rerankers trained with
retrieval-only supervision fail on real LoCoMo. v2 occupies the complementary
region v1 \S7 left open: it uses a pretrained cross-encoder backbone and
supervised (gold listwise) fine-tuning rather than a from-scratch architecture.
v2 is evidence that the missing ingredient in v1 \S7 was the supervised /
pretrained signal source, not the cascade idea.

\paragraph{v1 \S10 (CCGE-LA on top did not match mxbai).} v1 reported that even
with the CCGE-LA editor, absolute LoCoMo MRR remained well below mxbai. v2 narrows
this: v1 alone still loses to mxbai\_top500 (MRR 0.5824 vs 0.6688); v2 closes
about 85\% of the FULL-MRR gap between v1 and mxbai\_top500 (0.6560 vs 0.6688),
while mxbai\_top500 stays ahead on FULL. On two raw-dense-hard slices, however, v2 \emph{exceeds}
mxbai\_top500 (\S\ref{sec:results}). v2 and CCGE-LA are two independent opt-in
extensions; v2 neither contains nor replaces CCGE-LA.

\paragraph{v1 \S11 (three future-work directions).} v1 listed multi-backbone
checkpoint distribution, an end-to-end agent benchmark, and broader CCGE-LA
training as future work. v2 addresses a separate direction; all three v1 future-work items remain open.

Table~\ref{tab:rel} summarizes these relationships.

\begin{table}[htbp]
\centering
\caption{How ConvMemory v2 relates to specific results in the v1 report. v2
composes with v1 rather than replacing it, and does not revive or revisit v1's
negative attribution result.}
\label{tab:rel}
\renewcommand{\arraystretch}{1.25}
\begin{tabular}{>{\raggedright\arraybackslash}p{0.13\textwidth} >{\raggedright\arraybackslash}p{0.37\textwidth} >{\raggedright\arraybackslash}p{0.40\textwidth}}
\toprule
v1 section & v1 claim / result & v2 relationship \\
\midrule
\S3.3 & v1 headline deliberately avoids an mxbai teacher to avoid conflation &
v2 headline is also gold-only (teacher weight $0.0$); discipline preserved \\
\S3.4 & v1 default runs no per-pair transformer forward over the pool &
v1 default still satisfies this; v2 opts into 10 cross-encoder pair scorings on the
protected top-10, not 500 \\
\S5 & learned temporal window is not load-bearing (significant but not temporally
specific) & unchanged; v2's mechanism is non-temporal \\
\S7 & from-scratch stream rerankers fail without a teacher / supervised signal &
v2 uses gold listwise supervision on v1's top-10, the complementary
teacher-supervised region \\
\S10 & CCGE-LA on top did not close the mxbai MRR gap &
v2 closes most of the FULL-MRR gap, with mxbai\_top500 ahead by 0.013 on FULL;
v2 and CCGE-LA are independent opt-in extensions \\
\S11 & three future-work directions (multi-backbone, agent benchmark, broader
CCGE-LA) & all still open; v2 is a separate, unanticipated extension \\
\bottomrule
\end{tabular}
\end{table}

\section{Related Work}
\label{sec:related}

We keep related work focused on the components that directly determine v2’s design, evaluation setup, and claim boundaries.

ConvMemory v2 is built on the \path{cross-encoder/ms-marco-MiniLM-L-6-v2}
cross-encoder\footnote{\url{https://huggingface.co/cross-encoder/ms-marco-MiniLM-L-6-v2}}
as its scoring backbone, the same family v1 used as its distillation teacher. As
a high-cost full-pool reference point we use
mxbai-rerank-large-v1\footnote{\url{https://huggingface.co/mixedbread-ai/mxbai-rerank-large-v1}}
applied over the dense top-500. The base ConvMemory v1 reranker uses an MPNet
sentence encoder \cite{mpnet2020} and is described in the v1 report
\cite{convmemory_v1}. All
evaluation is on the LoCoMo conversational memory benchmark \cite{locomo2024}.
Cross-encoder reranking over a candidate prefix, and recall-preserving cascades
that reorder a protected set without changing recall, are standard IR patterns
that v2 composes for conversational memory retrieval.

\section{ConvMemory v2 Architecture}
\label{sec:arch}
\subsection{Recall-preserving cascade}
v2 is the last stage of a cascade whose earlier stages are unchanged from v1:
\begin{verbatim}
  query + memories
    --> dense MPNet candidate generation
    --> ConvMemory v1 rerank over the top-500 pool
    --> preserve the EXACT v1 top-10 set            <-- recall frozen here
    --> v2 cross-encoder reorders only those 10
    --> append v1's unchanged tail (ranks 11..N)
\end{verbatim}

\begin{figure}[htbp]
\centering
\includegraphics[width=\textwidth]{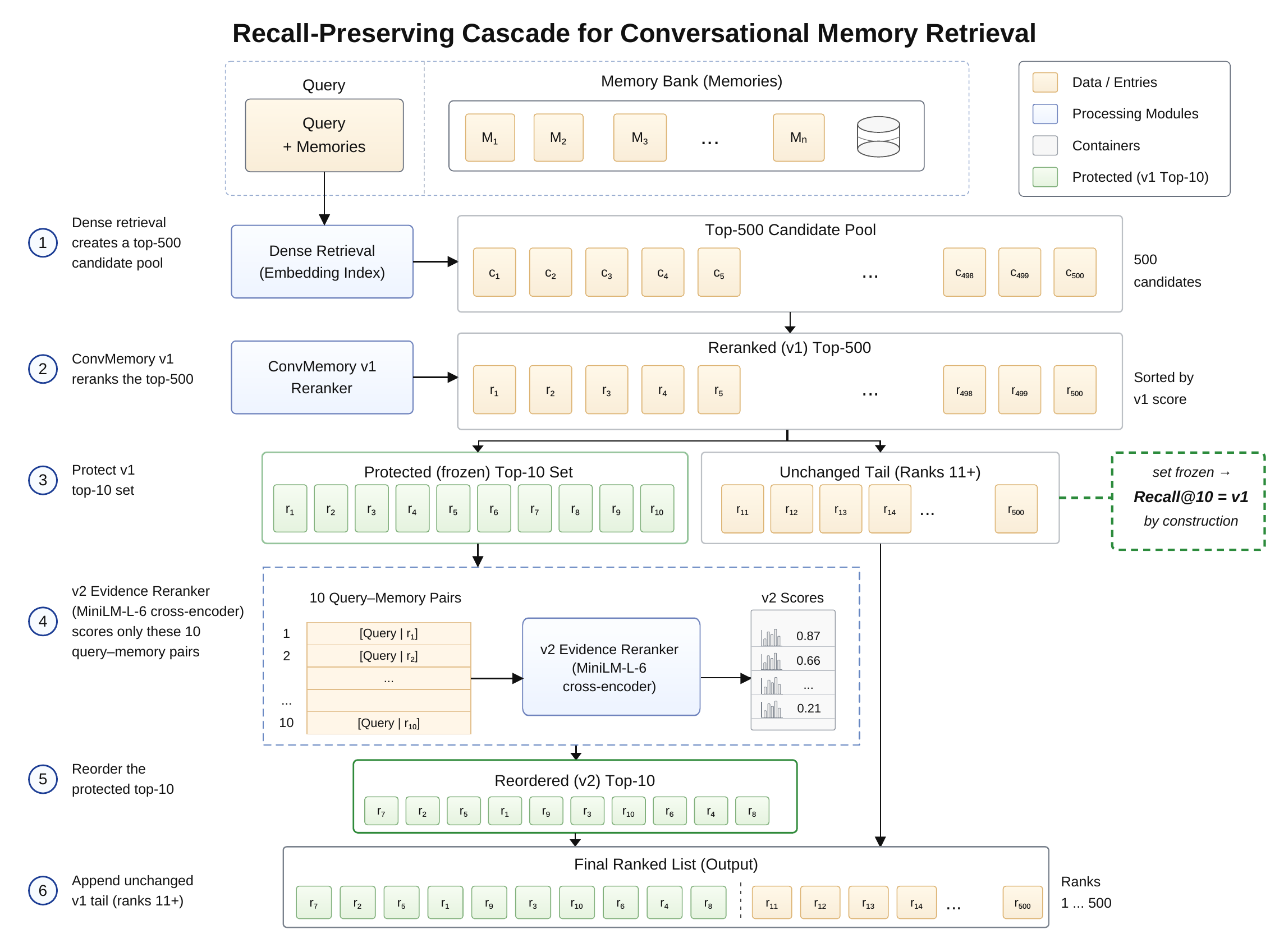}
\caption{The recall-preserving cascade in full. ConvMemory v1 reranks the dense
top-500; v2 reorders only the protected v1 top-10 (one pair scoring per protected candidate) and appends v1's unchanged tail. Because the top-10 set is
preserved, Recall@10 and Hit@10 equal v1's by construction.}
\label{fig:pipeline}
\end{figure}

Figure~\ref{fig:pipeline} shows the same cascade in full detail, including the
per-pair scoring of the protected top-10 and how the reordered set is recombined
with the unchanged tail.

The deployment rule is: take v1's top-10 set, score the ten (query, memory) text
pairs with the v2 cross-encoder, sort those ten by v2 score, and leave everything
at rank 11 and beyond exactly as v1 ordered it. Because the top-10 \emph{set} is
never changed, Recall@10 and Hit@10 are exactly equal to v1's --- this is a
constructive (by-design) property, not a statistical near-equality. The released
checkpoint and its provenance are described in \S\ref{sec:training}; the
isolation contract that constrains what v2 may read at inference is described in
\S\ref{sec:anticheat}.
\subsection{Scoring model and parameter count}
The v2 module is a fine-tuned \texttt{ms-marco-MiniLM-L-6-v2} cross-encoder. Its
parameter count, measured from the released checkpoint, is exactly
22{,}713{,}601 parameters (we write ``approximately 22.7M'' where a round figure
is convenient, but the measured value is 22{,}713{,}601). For comparison, the
lightweight v1 reranker is roughly 3.6M parameters and performs zero cross-encoder pair scorings per query; v2 adds a second, heavier stage that runs ten cross-encoder pair scorings per query on the protected prefix. Every cost statement in this report
distinguishes these two regimes (\S\ref{sec:cost}).
\subsection{Inference prompt format}
Each candidate is scored as a (query-side, memory-side) text pair using a
time-annotated format. With candidate positions $\{p_i\}$ and
$\texttt{max\_pos} = \max_i p_i$, the query side is rendered as
\begin{quote}
\texttt{QUERY\_TIME: \{max\_pos:.0f\}. \{query\}}
\end{quote}
and each candidate side as
\begin{quote}
\texttt{MEMORY\_TIME: \{pos:.0f\}. \{text\}}
\end{quote}
The position metadata is optional; when present it is the only temporal signal v2
sees, and the load-bearing ablation (\S\ref{sec:ablation}) shows it is not what
drives the gain --- candidate-specific memory text is.

\subsection{Inference procedure}

Algorithm~\ref{alg:v2} states the full v2 inference path. The key invariant is on
the last line: the returned top-10 \emph{set} equals v1's, so Recall@10 and
Hit@10 are preserved by construction.

\begin{algorithm}[htbp]
\caption{ConvMemory v2 inference (recall-preserving top-10 reorder)}
\label{alg:v2}
\begin{algorithmic}[1]
\Require query $q$; memory store $M$; base ConvMemory v1; attached evidence reranker $R$
\Ensure ranked memory list
\State $C \gets$ dense vector search over $M$, take dense top-500
\State $L \gets$ ConvMemory v1 rerank of $C$ \Comment{the v1 path; no per-pair cross-encoder scoring}
\State $P \gets \mathrm{first}_{10}(L)$ \Comment{protected top-10 set (inclusive)}
\State $T \gets$ remainder of $L$ after the first 10 items \Comment{unchanged tail}
\State $p_i \gets \mathrm{position}(m_i)$ for each $m_i \in P$
\State $\mathit{max\_pos} \gets \max_{m_i \in P} p_i$
\For{each $m_i \in P$}
    \State $q_i \gets$ \texttt{"QUERY\_TIME:\ \{max\_pos\}.\ "} $+\, q$
    \State $d_i \gets$ \texttt{"MEMORY\_TIME:\ \{}$p_i$\texttt{\}.\ "} $+\, \mathrm{text}(m_i)$
    \State $s_i \gets R.\mathrm{score}(q_i, d_i)$ \Comment{one pair scoring}
\EndFor
\State $P' \gets$ sort $P$ by $s_i$ descending
\State \Return $P' \,\|\, T$
\end{algorithmic}
\vspace{2pt}
\noindent\textit{Invariant:} $\mathrm{set}(P') = \mathrm{set}(P)$, so Recall@10 and Hit@10
equal v1's by construction. The reranker performs exactly $|P|=10$ pair scorings per query, never $|C|=500$.
\end{algorithm}

\section{Training}
\label{sec:training}

\subsection{Objective}

v2 is trained with a listwise gold-only objective over the protected top-10. For
a question with candidate logits $z_i$ and a gold indicator $g_i$ (normalized to
sum to one over the gold memories), the loss is the listwise cross-entropy

\[
\mathcal{L}_{\text{gold}} = -\sum_i g_i \, \log\!\big(\mathrm{softmax}(z)_i\big),
\]

with the softmax taken over the ten protected candidates. An optional
cross-encoder teacher term (a KL-style soft-label loss against an mxbai soft
teacher) is supported, giving the general objective
$\mathcal{L} = w_{\text{gold}}\,\mathcal{L}_{\text{gold}} + w_{\text{teacher}}\,\mathcal{L}_{\text{teacher}}$.
\textbf{The headline arm is gold-only}: $w_{\text{gold}} = 1.0$,
$w_{\text{teacher}} = 0.0$. A separate cross-encoder-teacher arm
($w_{\text{teacher}} = 0.25$) was run for comparison and reaches FULL MRR 0.6546,
within noise of the gold-only 0.6560 (\S\ref{sec:results}); the teacher term is
therefore not load-bearing and is not part of the headline.

\paragraph{Questions whose gold is outside the protected top-10.} v1's top-10
recall is 0.7798, so for roughly 22\% of training questions no gold memory is
present in the protected top-10. These questions still pass through the forward
pass, but their gold target is replaced by a uniform placeholder and a
\texttt{has\_gold} mask zeroes their contribution to the listwise loss; since the
headline arm also uses teacher weight $0.0$, such questions contribute nothing
to the training signal. This is consistent with the recall-preserving contract:
v2 cannot recover a gold item that v1 failed to place in the protected set, so
these questions carry no usable ordering signal. Evaluation, in contrast, is
over \emph{all} test questions, including those v2 cannot recover, so the
headline MRR already prices in the recall ceiling.

\subsection{Hyperparameters and split}

Training uses AdamW with learning rate $2\times10^{-5}$, weight decay $0.01$,
batch size $8$, one epoch, and a linear warmup capped at 100 steps. The training
data is the LoCoMo dev split with \texttt{dev\_ratio} $=0.5$, where the split is
taken by conversation id (\path{question_id.split("::",1)[0]}) so that dev and
test conversations are disjoint (\S\ref{sec:anticheat}). All headline numbers are
5-seed (seeds 7, 11, 23, 31, 47). The cross-encoder backbone is
\path{cross-encoder/ms-marco-MiniLM-L-6-v2} with \texttt{max\_length} $=256$ and
\texttt{top\_k} $=10$.

\subsection{Training source and released-checkpoint provenance}

The exact internal reproduction of the v363 5-seed method-level headline numbers
uses the internal experiment script
\path{experiments/v361_top10_evidence_reranker.py}; this is the script to
cite for the headline numbers. A separate public, general-purpose training entry
point is provided as \path{examples/train_evidence_reranker.py}; it is the
user-facing recipe for training an evidence reranker on user records and is not
the exact LoCoMo-locked harness.

The released checkpoint requires a careful provenance statement. The original
v361 5-seed run did not save its per-seed checkpoints. The checkpoint published
on the Hugging Face Hub
(\path{Purdy0228/ConvMemory-v2-Evidence-Reranker}) is a \textbf{seed-7
representative checkpoint}, exported from the same v361 gold-only recipe after
v0.5.0 packaging (provenance:
\path{results/v365_v05_evidence_reranker_}\path{checkpoint/seed_7/MANIFEST.json},
which records a teacher weight of $0.0$, a training target of gold-only listwise
retrieval cross-entropy, and a source experiment of
\path{v361_top10_evidence_reranker.py}). Consequently the headline FULL MRR of 0.6560
is a \textbf{method-level 5-seed estimate}, not a separately measured score of
this exact single checkpoint. We keep this distinction explicit throughout
(\S\ref{sec:results}, \S\ref{sec:limitations}).

\section{Experimental Protocol}
\label{sec:protocol}

\paragraph{Dataset and task.} All experiments are on the LoCoMo conversational
memory retrieval setting \cite{locomo2024}: given a query against an accumulated
multi-session conversation, retrieve the gold memory (or memories) supporting the
answer. We evaluate at the retrieval stage only, without a downstream answer generator.

\paragraph{Candidate pool.} The candidate pool is the dense MPNet top-500, the
same pool used for the v1, v2, and mxbai comparisons where applicable. v2 only
ever sees, and only ever reorders, the protected v1 top-10 drawn from this pool.

\paragraph{v1 anchor.} The v1 baseline is the paper-compatible ConvMemory
candidate-local path reproduced in v359, matching the v1 report's v0.40 Table-3
style. Its 5-seed FULL metrics are R@10 0.7798, MRR 0.5824, H@1 0.4440, which
match the v1 report and anchor every v2$-$v1 delta.

\paragraph{Seeds and split.} All headline numbers are 5-seed (seeds 7, 11, 23,
31, 47). The train/test split is produced by \texttt{choose\_split} on the
conversation (sample) id, where
\path{sample_id = question_id.split("::",1)[0]} and
\texttt{dev\_ratio} $= 0.5$. The split is taken over distinct sample ids and then
materialized to examples, so dev (training) and test conversations are disjoint
within each seed. Each seed thus induces its own conversation-level split: the seed is passed into \texttt{choose\_split}, which shuffles the conversation ids with
\texttt{random.Random(seed)} before partitioning, so seeds vary the dev/test
partition as well as model initialization. The FULL row count $n=4955$ is the
\emph{total} number of test questions pooled across the five seed-specific test
sets (per-seed sizes 938, 1135, 937, 990, 955). Metrics are computed per
(method, seed) and then averaged over seeds; paired bootstrap intervals resample
paired per-question differences over the pooled rows, with the resampling unit
keyed by \texttt{seed::split::question\_id}. Because test rows are pooled across seed-specific splits, the bootstrap interval should be read as a paired method-level uncertainty estimate over evaluated seed--question instances, not as a confidence interval over a single fixed held-out test set.

\paragraph{Metrics.} We report Recall@10, MRR, and H@1. Because v2 preserves the
v1 top-10 set, its Recall@10 (and Hit@10) equal v1's exactly --- this is a
by-design equality, not an empirical near-tie, so v2$-$v1 deltas on R@10 are an
exact zero. The metric that moves is ordering quality inside the protected set
(MRR, H@1).

\paragraph{Significance.} Confidence intervals are from a question-level paired
bootstrap over the pooled per-seed test rows ($n=4955$ on FULL, pooled across five seeds), resampling questions with replacement and recomputing the paired metric difference. We report 95\% intervals; a delta is called significant when its interval excludes zero.

\paragraph{Hard slices.} Beyond FULL we report three diagnostic slices:
\begin{itemize}
    \item \textbf{T\_SUP\_auto}: supersession-style questions where a later
    memory supersedes an earlier one. This slice is assigned \emph{automatically}
    by heuristic, not human-audited; we treat it as an automatic slice, and its
    labels are heuristic rather than human-audited.
    \item \textbf{RAW\_TOP1\_WRONG\_GOLD\_IN\_POOL}: questions where the raw dense
    top-1 is wrong but the gold memory is nonetheless somewhere in the pool ---
    i.e.\ reranking has something to recover.
    \item \textbf{RAW\_RESCUABLE\_STALE\_TOP1}: questions where the raw dense
    top-1 is a stale memory but a rescuable correct memory is present.
\end{itemize}
On both raw-dense-hard slices, v1's protected top-10 has high recall
(R@10 $\approx 0.93$), which is what gives the recall-preserving cascade room to
beat a full-pool cross-encoder on those slices (\S\ref{sec:results}).

\section{Inference Isolation and Shortcut Controls}
\label{sec:anticheat}
v2's inference contract is deliberately narrow; we refer to it as the
anti-shortcut contract, following the v1 report. The only inputs the scorer
accepts are: query text; candidate memory id and memory text; optional candidate
position/time metadata; and the protected v1 top-10 candidate set. Gold labels
and teacher (mxbai / cross-encoder) scores are used only as training or
evaluation targets and never as inference features.

The public API enforces this with a fixed \texttt{FORBIDDEN\_FIELDS} set. If any
candidate passed at inference contains any of the following keys, the API raises
\texttt{ValueError}:

\begin{quote}
\texttt{gold}, \texttt{gold\_ids}, \texttt{is\_current}, \texttt{is\_latest},
\texttt{is\_stale}, \texttt{stale}, \texttt{answer}, \texttt{answer\_text},
\texttt{ce\_score}, \texttt{mxbai\_score}, \texttt{teacher\_score},
\texttt{gpt\_label}, \texttt{entity\_id}, \texttt{slot\_id}.
\end{quote}

Two properties are checked mechanically rather than asserted in prose. The test
\path{test_evidence_reranker_rejects_forbidden_fields} iterates over every
forbidden field and confirms the API rejects it. The test
\path{test_default_behavior_unchanged} confirms that the default (v1) path is
byte-identical whether or not the v2 module is present, so enabling v2 is a true
opt-in. A third test confirms recall preservation: the returned top-10 set is
identical with and without v2.

Train/test isolation is by conversation: the split function partitions on
\path{sample_id(question_id) = question_id.split("::",1)[0]}, so that no
conversation appears in both the dev (training) and test partitions for a given
seed. The final v361 headline is trained without GPT-labeled data; earlier GPT
experiments fall outside the v2 claim.

Table~\ref{tab:audit} collects the isolation contract as an audit checklist.

\begin{table}[htbp]
\centering
\caption{Anti-shortcut / isolation audit checklist. Each property is backed by a
named mechanism or test rather than by prose assertion.}
\label{tab:audit}
\renewcommand{\arraystretch}{1.2}
\begin{tabular}{>{\raggedright\arraybackslash}p{0.30\textwidth} >{\raggedright\arraybackslash}p{0.62\textwidth}}
\toprule
Aspect & Mechanism \\
\midrule
Inference accepts & query text; candidate id; candidate text; optional
time/position; protected v1 top-10 \\
Inference rejects & \texttt{gold}, \texttt{gold\_ids}, \texttt{is\_current},
\texttt{is\_latest}, \texttt{is\_stale}, \texttt{stale}, \texttt{answer},
\texttt{answer\_text}, \texttt{ce\_score}, \texttt{mxbai\_score},
\texttt{teacher\_score}, \texttt{gpt\_label}, \texttt{entity\_id},
\texttt{slot\_id} (\texttt{ValueError}) \\
Forbidden-field test & \path{test_evidence_reranker_rejects_forbidden_fields} \\
Opt-in / no-regression test & \path{test_default_behavior_unchanged} (default
path byte-identical) \\
Recall-preservation test & \path{test_evidence_reranker_recall_preserving} \\
Save/load round-trip test & \path{test_evidence_reranker_save_load_roundtrip} \\
Train/test isolation & conversation-level \texttt{sample\_id} split,
\texttt{dev\_ratio} $=0.5$ \\
Gold use & training target only, never an inference feature \\
Teacher (mxbai/CE) use & optional training/evaluation target only; headline uses
teacher weight $0.0$ \\
\bottomrule
\end{tabular}
\end{table}

\section{Main Results}
\label{sec:results}

\subsection{Headline (FULL) and slice results}

Table~\ref{tab:headline} reports the v363 verifier-packet headline numbers
(5-seed method-level; $n=4955$ pooled test rows on FULL). v2 is the
\texttt{v361\_top10\_gold\_listwise} arm. R@10 is constant across the v1-derived
rows by construction (the protected top-10 set is preserved).

\begin{table}[htbp]
\centering
\caption{LoCoMo FULL results, 5-seed method-level ($n=4955$). v2 preserves v1's
Recall@10 by construction and improves MRR and H@1. mxbai\_ce\_top500 is a
high-cost full-pool cross-encoder reference (not an upper bound);
mxbai\_top10\_guard is mxbai restricted to the same protected top-10
(set-matched control, i.e.\ same ten candidates and same ten pair scorings, but
not the same wall-clock cost); oracle\_top10\_guard is the true ceiling for any reordering of v1's
top-10.}
\label{tab:headline}
\begin{tabular}{lrrr}
\toprule
Arm & R@10 & MRR & H@1 \\
\midrule
raw\_dense                          & 0.5345 & 0.3254 & 0.1937 \\
ConvMemory v1 (paper-compatible)    & 0.7798 & 0.5824 & 0.4440 \\
\textbf{v2 (v361 top10 gold\_listwise)} & 0.7798 & \textbf{0.6560} & \textbf{0.5474} \\
mxbai\_ce\_top500 (strong full-pool CE ref) & 0.8080 & 0.6688 & 0.5646 \\
mxbai\_top10\_guard (set-matched)   & 0.7798 & 0.6623 & 0.5598 \\
oracle\_top10\_guard (true ceiling) & 0.7798 & 0.8415 & 0.8350 \\
\bottomrule
\end{tabular}
\end{table}

Paired bootstrap of v2 minus v1 on the full set ($n=4955$):

\begin{itemize}
    \item MRR delta $+0.0734$, 95\% CI $[+0.0645, +0.0827]$.
    \item H@1 delta $+0.1033$, 95\% CI $[+0.0906, +0.1171]$.
    \item R@10 delta $= 0$ (constructive zero; the top-10 set is unchanged).
\end{itemize}

\paragraph{Naming discipline for the mxbai reference.} We treat mxbai\_ce\_top500 as a strong but expensive full-pool cross-encoder
reference rather than an upper bound or ceiling. The true ceiling for any method that may only
reorder v1's protected top-10 is \texttt{oracle\_top10\_guard} (FULL MRR 0.8415),
which leaves substantial headroom above v2.

\subsection{Hard slices: where v2 exceeds the full-pool reference}

Table~\ref{tab:slices} reports two raw-dense-hard slices alongside the
supersession slice (5-seed means). On both hard slices, v2 \emph{exceeds}
mxbai\_top500.

\begin{table}[htbp]
\centering
\caption{Slice MRR (5-seed mean). v2$-$v1 deltas are paired-bootstrap with 95\%
CI. On RAW\_TOP1\_WRONG and RAW\_RESCUABLE\_STALE, v2
exceeds mxbai\_top500; on T\_SUP\_auto and on FULL it does not. Slice names are
abbreviated: RAW\_TOP1\_WRONG is RAW\_TOP1\_WRONG\_GOLD\_IN\_POOL (raw dense top-1
wrong but gold in pool); RAW\_RESCUABLE\_STALE is RAW\_RESCUABLE\_STALE\_TOP1 (raw
dense top-1 stale but a rescuable correct memory present).}
\label{tab:slices}
{\setlength{\tabcolsep}{3pt}
\begin{tabular}{lrrrrr}
\toprule
Slice & $n$ & v1 MRR & v2 MRR & v2$-$v1 (95\% CI) & mxbai\_top500 MRR \\
\midrule
T\_SUP\_auto          & 708  & 0.5649 & 0.6469 & $+0.082$\,[+.058,+.106] & 0.6572 \\
RAW\_TOP1\_WRONG       & 3183 & 0.6344 & 0.7486 & $+0.113$\,[+.101,+.125] & 0.6970 \\
RAW\_RESCUABLE\_STALE  & 1823 & 0.6235 & 0.7510 & $+0.125$\,[+.110,+.142] & 0.6857 \\
\bottomrule
\end{tabular}
}
\end{table}

\paragraph{Why v2 beats mxbai\_top500 on these slices.} The mechanism is a recall asymmetry in the protected pool; it reflects pool
composition rather than v2's scorer being stronger than mxbai in general. On these two raw-dense-difficult slices, v1's protected
top-10 has R@10 $\approx 0.93$, higher than mxbai's own top-10 recall when mxbai
reranks the full top-500 (R@10 $\approx 0.85$ on these slices);
Table~\ref{tab:asym} reports the exact per-slice values. So
``v1's high-recall protected pool $+$ v2's precise reordering'' can outscore
``mxbai reranking a top-500 whose own top-10 already lost more gold.'' This is a
genuine selling point of the recall-preserving design, but it is slice-specific:
on FULL MRR and on T\_SUP\_auto, mxbai\_top500 is still ahead.

\begin{table}[htbp]
\centering
\caption{The recall asymmetry behind Table~\ref{tab:slices}. R@10(v1 protected)
is the recall of v1's protected top-10 on the slice; R@10(mxbai top-10) is the
recall of mxbai\_top500's own top-10 after reranking the full pool. On the two
raw-dense-hard slices v1's protected set retains more gold; on T\_SUP\_auto the
asymmetry is reversed, consistent with v2 trailing mxbai\_top500 there.}
\label{tab:asym}
\begin{tabular}{lrr}
\toprule
Slice & R@10 (v1 protected) & R@10 (mxbai top-10) \\
\midrule
T\_SUP\_auto          & 0.759 & 0.790 \\
RAW\_TOP1\_WRONG       & 0.931 & 0.854 \\
RAW\_RESCUABLE\_STALE  & 0.929 & 0.840 \\
\bottomrule
\end{tabular}
\end{table}

\paragraph{Scope of these results.} These results are LoCoMo-only and
retrieval-stage. They are evidence for slice-specific gains under the
recall-preserving cascade, not for v2 surpassing mxbai in general (it does not on
FULL or T\_SUP), for generalization beyond conversational memory, or for
downstream answer-quality gains in an end-to-end agent/QA pipeline (not evaluated
here). v2's gains are
upper-bounded by v1's top-10 recall: any gold memory v1 fails to place in the
protected top-10 is unrecoverable by v2.

\subsection{The top-10 reordering ceiling}
\label{sec:ceiling}

To put v2's FULL MRR in context, Table~\ref{tab:ceiling} places it between the two
set-matched references that operate on the same protected top-10.

\begin{table}[htbp]
\centering
\caption{FULL MRR on the protected top-10. \texttt{mxbai\_top10\_guard} is mxbai
restricted to the same ten candidates (set-matched: same candidates and same
number of pair scorings, not the same cost);
\texttt{oracle\_top10\_guard} is the true ceiling for any reordering of v1's
top-10 (it places the gold first whenever the gold is in the set). v2 recovers a
large fraction of the v1$\to$mxbai\_top10\_guard headroom but leaves substantial
room below the oracle.}
\label{tab:ceiling}
\begin{tabular}{lr}
\toprule
Arm (all reorder the same v1 top-10) & FULL MRR \\
\midrule
ConvMemory v1                       & 0.5824 \\
v2 (v361 gold\_listwise)            & 0.6560 \\
mxbai\_top10\_guard (set-matched)   & 0.6623 \\
oracle\_top10\_guard (true ceiling) & 0.8415 \\
\bottomrule
\end{tabular}
\end{table}

Two points follow. First, on the set-matched comparison (same ten candidates),
v2 (0.6560) is close to mxbai\_top10\_guard (0.6623), and v2's MiniLM-L-6 backbone
is $6.7\times$ cheaper to run on that set (\S\ref{sec:cost}). Second, the true
ceiling for reordering v1's top-10 is oracle\_top10\_guard at 0.8415, far above
all learned arms; the headroom that remains is bounded by ordering quality, while
the headroom that is permanently lost is bounded by v1's top-10 recall.

\section{Load-Bearing Ablation}
\label{sec:ablation}

To identify what actually drives v2's MRR gain, the v364 audit retrains the
v2-style full-text arm in the same harness as four perturbation arms, all 5-seed,
all preserving the exact v1 top-10 set (so all have R@10 $=0.7798$ on FULL).
Table~\ref{tab:ablation} reports FULL MRR.

The four perturbation arms are designed to remove or corrupt exactly one source
of signal each, while keeping the protected top-10 set (and hence Recall@10)
fixed:
\begin{itemize}
    \item \texttt{no\_memory\_text}: the candidate memory text is removed
    entirely, leaving only the query side and any positional prefix.
    \item \texttt{random\_other\_query\_text} (the name is historical; it is the
    candidate \emph{memory} text, not the query, that is replaced): each
    candidate's text is replaced with text drawn from \emph{other} questions ---
    the cross-question stress test, which breaks query--candidate alignment
    while keeping superficially plausible text.
    \item \path{shuffled_memory_text}: the candidate texts are permuted
    \emph{within the same question} --- the same-topic-but-wrong-candidate stress
    test, which keeps the texts on-topic but attaches them to the wrong
    candidate.
    \item \texttt{scalar\_only}: only rank/score/time metadata is kept; the
    memory text is withheld, isolating whatever a non-textual shortcut could
    achieve.
\end{itemize}

\begin{table}[htbp]
\centering
\caption{Token-evidence load-bearing ablation, FULL MRR, 5-seed. All arms preserve
v1's top-10 set. The within-harness full-text baseline is 0.6677 (see
\S\ref{sec:tworuns} on why this differs slightly from the 0.6560 headline). The
three text perturbations all fall below raw dense (0.3254).}
\label{tab:ablation}
\begin{tabular}{lr}
\toprule
Arm & FULL MRR \\
\midrule
v361\_full\_text (within-harness baseline) & 0.6677 \\
v364\_no\_memory\_text                     & 0.2966 \\
v364\_random\_other\_query\_text           & 0.2506 \\
v364\_shuffled\_memory\_text               & 0.2731 \\
v364\_scalar\_only                         & 0.5792 \\
\bottomrule
\end{tabular}
\end{table}

\begin{figure}[htbp]
\centering
\includegraphics[width=0.9\textwidth]{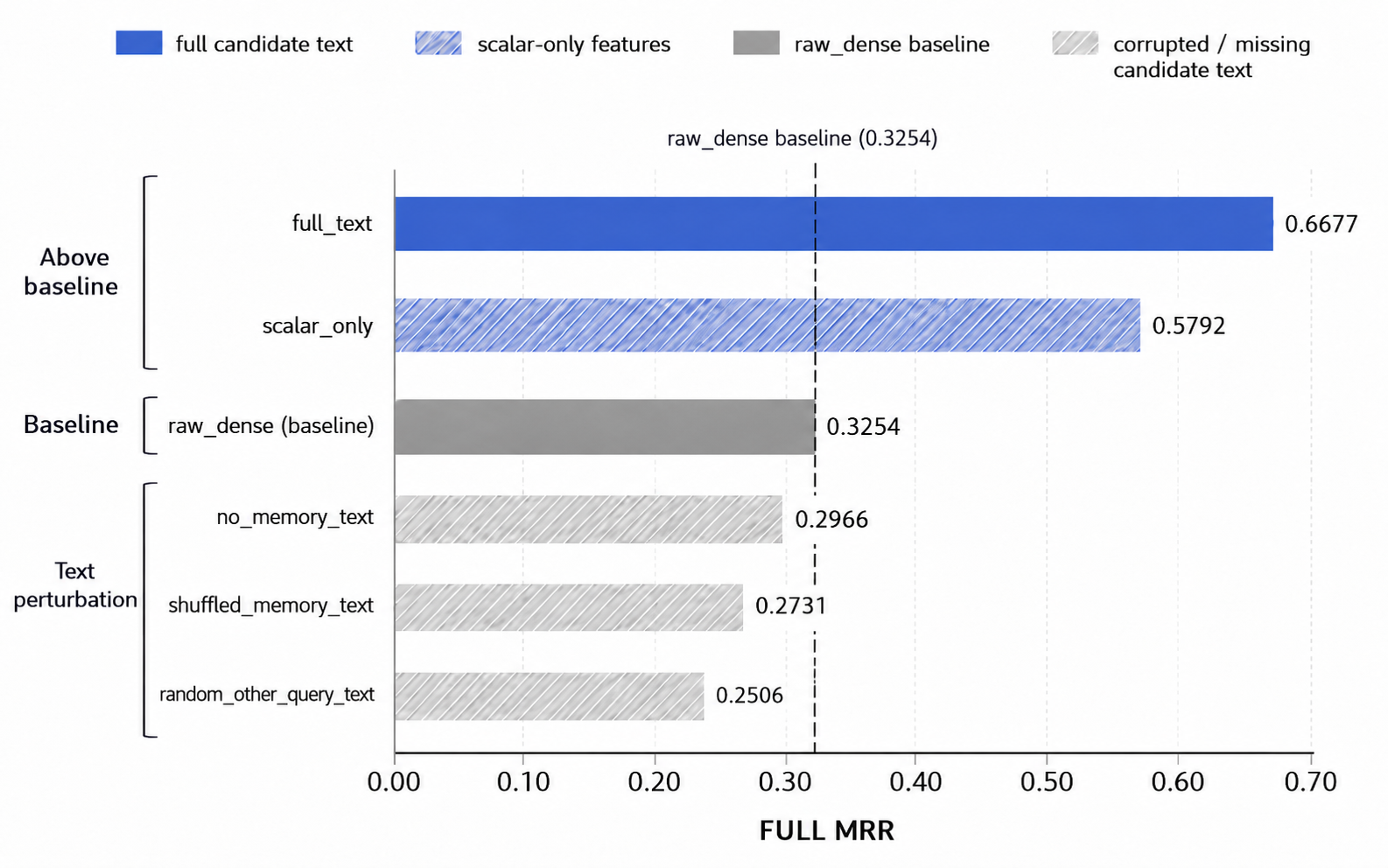}
\caption{Token-evidence load-bearing ablation (FULL MRR, 5-seed). All arms
preserve v1's top-10 set. The three text-perturbation arms (no\_memory\_text,
shuffled\_memory\_text, random\_other\_query\_text) all fall below the raw\_dense
baseline (0.3254, dashed line): corrupting candidate-specific memory text does
not merely erase the gain, it scores below the unreranked baseline.
scalar\_only stays near v1 level.}
\label{fig:ablation}
\end{figure}

Figure~\ref{fig:ablation} visualizes the same ablation; the three
text-perturbation arms all fall below the raw-dense baseline, while
\texttt{scalar\_only} stays near v1 level.

Paired bootstrap, full text minus each ablation on FULL MRR (all 95\% CIs exclude
zero):
\begin{itemize}
    \item full $-$ no\_memory\_text: $+0.3712$ $[+0.3599, +0.3829]$.
    \item full $-$ random\_other\_query\_text: $+0.4173$ $[+0.4067, +0.4284]$.
    \item full $-$ shuffled\_memory\_text: $+0.3948$ $[+0.3834, +0.4060]$.
    \item full $-$ scalar\_only: $+0.0881$ $[+0.0801, +0.0969]$.
\end{itemize}
\paragraph{Key finding.} The three text perturbations
($0.2506$, $0.2731$, $0.2966$) all fall \emph{below} raw dense retrieval
($0.3254$). This is the cleanest fingerprint that token-on-memory-text alignment
is load-bearing: when the candidate-specific text is removed, shuffled within the
question, or swapped for text from other questions, the model does not gracefully
fall back to a sensible default --- it is confidently wrong, scoring below the
unreranked dense order. By contrast, the scalar-only arm (rank/time/score
features without memory text) stays at roughly v1 level (0.5792 vs v1's 0.5824),
showing that scalar shortcuts alone cannot reproduce the gain. The mechanism is
candidate-specific text interaction, not rank/time/score shortcuts.
The two text stress tests probe complementary failure modes and both collapse.
\path{random_other_query_text} (0.2506) confirms the model is not exploiting
generic, query-independent text statistics: when candidate texts come from
unrelated questions, scoring is worse than dense order. \path{shuffled_memory_text}
(0.2731) is the harder, same-question control: the texts are exactly the
candidate texts for that question, merely reattached to the wrong candidates, and
the model still collapses --- so it is the \emph{alignment} between a specific
candidate's text and the query, not mere topical presence of plausible text, that
carries the signal. That all three text-damaged arms land below raw dense
(0.3254) rather than merely below full text is the strong form of the result: a
model that had only learned a mild text prior would degrade toward the dense
baseline, not below it. Scoring below dense means the corrupted text is actively
misleading the reranker, which strongly supports the interpretation that the uncorrupted candidate-specific text was load-bearing.

\subsection{The two v2 numbers (0.6560 vs 0.6677)}
\label{sec:tworuns}

There are two method-level v2 MRR estimates in this report, and we keep them
distinct. The canonical headline is the v361 5-seed method-level run: FULL MRR
$0.5824 \to 0.6560$. The released HF checkpoint
(\path{Purdy0228/ConvMemory-v2-Evidence-Reranker}) is a seed-7 representative
checkpoint exported from the same v361 gold-only recipe after v0.5.0 packaging;
the $0.6560$ figure is the method-level 5-seed estimate, \textbf{not} a separately
measured score of this exact single checkpoint. The v364 ablation-harness baseline
is $0.6677$ (full\_text retrained inside the v364 ablation script as a fresh
5-seed run, for apples-to-apples comparison with the four perturbation arms). The
$+0.012$ gap between $0.6560$ and $0.6677$ is about $1.3\sigma$ of fresh-training
seed variance (MRR std $\approx 0.009$); both method-level estimates are
significantly above v1, with bootstrap CIs that do not cross zero.

\section{Cost}
\label{sec:cost}

\begin{figure}[htbp]
\centering
\includegraphics[width=0.85\textwidth]{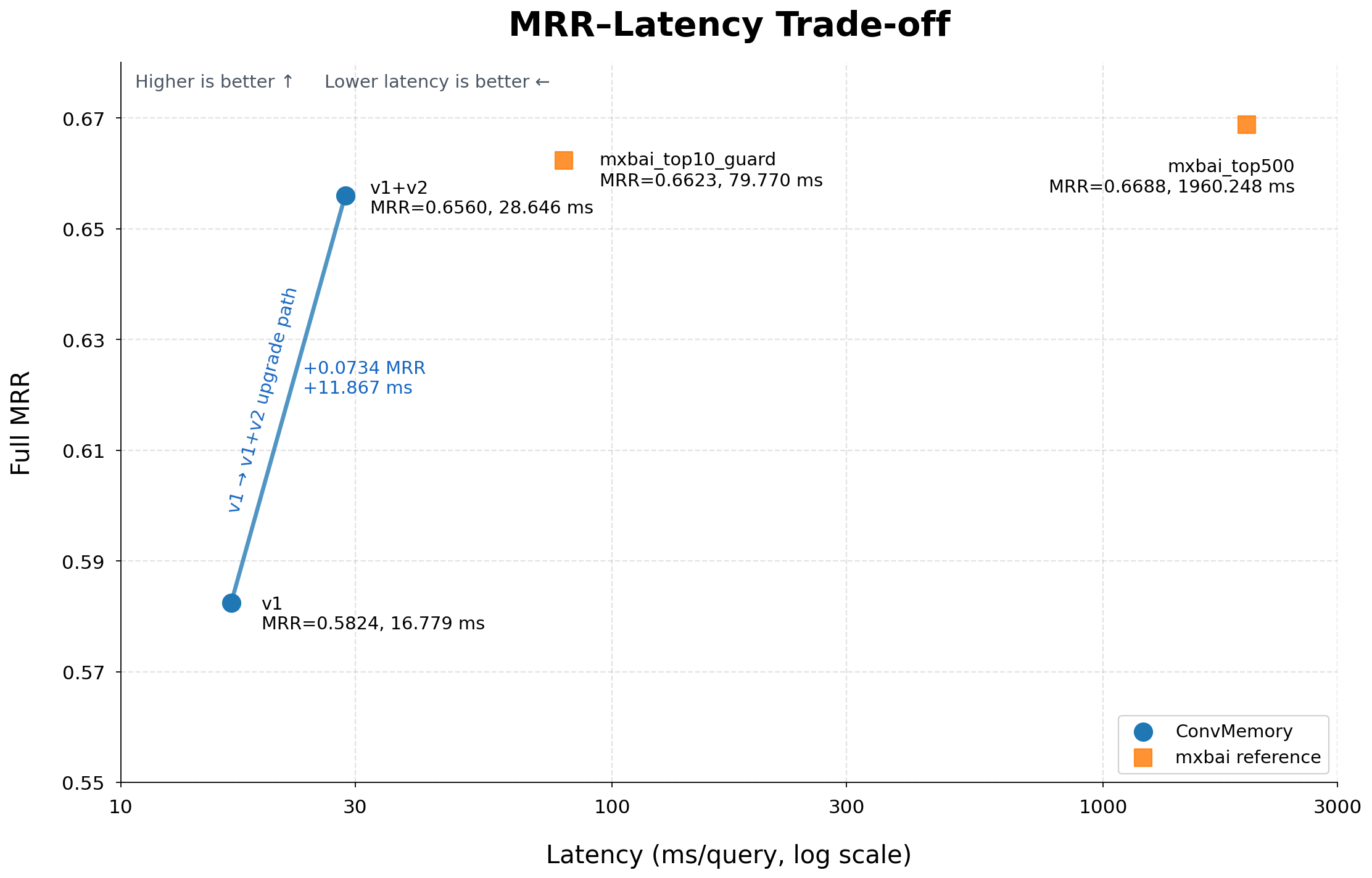}
\caption{MRR--latency trade-off on LoCoMo (x-axis: log scale). v1+v2 recovers
most of the MRR headroom above v1 at $\sim$1.7$\times$ v1's latency, and stays
within 0.013 FULL MRR of the high-cost full-pool reference mxbai\_top500 while
running about 68$\times$ cheaper. mxbai\_top500 remains the most accurate but by
far the slowest, and remains ahead of v2 on FULL MRR.}
\label{fig:frontier}
\end{figure}

Figure~\ref{fig:frontier} plots the cost--quality trade-off across the four
reranking paths; the precise latency figures follow in Table~\ref{tab:cost}.

Table~\ref{tab:cost} reports the v362 latency probe (200 timed queries after
warmup, RTX 4080 SUPER). The full protocol is in Appendix~\ref{app:latency}.
\begin{table}[htbp]
\centering
\caption{Latency (RTX 4080 SUPER, 200 timed queries after warmup). ``pair scorings'' is the number of cross-encoder query--memory pair evaluations per query. v1 runs none;
the v2 evidence stage runs ten (one per protected candidate); a full-pool
cross-encoder runs 500.}
\label{tab:cost}
\renewcommand{\arraystretch}{1.15}
\begin{tabular}{>{\raggedright\arraybackslash}p{0.24\textwidth} r r r >{\raggedright\arraybackslash}p{0.30\textwidth}}
\toprule
Component & Cands & pair scorings & ms/q & Comment \\
\midrule
v1 top500            & 500 & 0   & 16.779   & default cheap path \\
v2 (v361 top10 only) & 10  & 10  & 11.867   & evidence stage only \\
v1 + v2              & 500 then 10 & 10 & 28.646 & deployed v2 path ($1.71\times$ v1) \\
mxbai\_top10\_guard  & 10  & 10  & 79.770   & same top-10, stronger/heavier CE \\
mxbai\_top500        & 500 & 500 & 1960.248 & high-cost full-pool reference \\
\bottomrule
\end{tabular}
\end{table}

\paragraph{What the latency figures include and exclude.} Each measured path
includes the model scoring it names: ConvMemory v1 scoring over the top-500 for
the v1 rows; the v2 cross-encoder scoring over the protected top-10 for the v2
rows; and mxbai cross-encoder scoring over the stated candidate count for the
mxbai rows. The figures exclude vector-database I/O, network latency, downstream
LLM answer generation, and prompt construction; they reflect reranking-stage
compute on the stated candidates. Embedding precomputation is excluded except
where it is intrinsically part of a measured path. These are single-configuration
indicative measurements, not benchmark-grade latency claims; the full protocol and
caveats are in Appendix~\ref{app:latency}.

\paragraph{Cost framing.} v1 alone preserves the v1 \S3.4 property (zero per-pair
cross-encoder scoring over the pool). v1+v2 adds a bounded precision stage: ten
cross-encoder pair scorings per query on the protected top-10, where the v2 module
itself is 22{,}713{,}601 parameters. This adds a new point on the cost--quality
frontier at roughly $1.7\times$ v1's latency and about $68\times$ cheaper than a
full-pool mxbai cross-encoder in this measurement; v1 remains the cheaper default path alongside it. Comparing the two cross-encoders on the
identical protected top-10 (set-matched), v2 (11.867 ms/q) is $6.72\times$ cheaper than mxbai
(79.770 ms/q), because v2's MiniLM-L-6 backbone is far smaller than mxbai-large.

\section{Limitations}
\label{sec:limitations}

\paragraph{Recall is capped by v1.} v2 only reorders v1's protected top-10, so it
inherits v1's top-10 recall as a hard ceiling. Any gold memory that v1 fails to
place in the top-10 is permanently unrecoverable by v2. The relevant ceiling is
\texttt{oracle\_top10\_guard} (FULL MRR 0.8415), which is the best any reorderer
of v1's top-10 could achieve; v2's 0.6560 leaves real headroom, and that headroom
is split between better ordering (recoverable) and missing recall (not).

\paragraph{LoCoMo-specific fine-tuning, no cross-domain evidence.} The headline is
a LoCoMo conversational-memory result. Transfer to other domains, other conversation distributions, or document
retrieval is untested here and outside the scope of this report's claims. A cross-domain user should retrain or at least re-validate their own
evidence reranker rather than assume the LoCoMo checkpoint transfers.

\paragraph{Single-hardware latency.} All cost numbers come from one consumer GPU
(RTX 4080 SUPER) in a single run of 200 timed queries. Data-center GPUs, smaller
GPUs, CPU inference, and kernel-fused servers would shift both the absolute
numbers and the ratios; practitioners should measure on their target hardware.

\paragraph{Retrieval-stage only.} Every metric here is retrieval-stage MRR / H@1 /
Recall@10. The evaluation stops at retrieval; whether v2's ordering gains translate into
downstream answer-quality gains in an end-to-end agent or QA pipeline is left to
future work.

\paragraph{Below mxbai on FULL.} On FULL MRR, v2 sits 0.013 below
mxbai\_top500 (0.6560 vs 0.6688), and on the T\_SUP\_auto slice 0.6469 vs 0.6572.
v2's advantage over mxbai\_top500 holds only on the two raw-dense-hard slices of
Table~\ref{tab:slices} and is a slice-specific recall-asymmetry effect rather than
general dominance.

\paragraph{Released checkpoint vs.\ headline number.} The published checkpoint is a
single seed-7 representative of the gold-only recipe; the headline figures are
5-seed method-level estimates (\S\ref{sec:tworuns}, Appendix~\ref{app:provenance}).
The two are not interchangeable, and downloading the single checkpoint should not
be expected to reproduce the 5-seed aggregate exactly. A cross-encoder-teacher
variant exists ($w_{\text{teacher}}=0.25$, FULL MRR 0.6546) but is not the
headline; the headline is gold-only listwise.

\paragraph{Not a replacement for other extensions.} v2 is an opt-in stage, not a
replacement for v1 (the cheaper default) nor for the separate CCGE-LA conflict
editor or the Memory-MLA recall expander; these are independent components, each composing with v2 as a separate opt-in stage.

\section{Reproducibility}
\label{sec:repro}

The package is installable via \texttt{pip install convmemory==0.5.0}. The
released v2 checkpoint is on the Hugging Face Hub at
\path{Purdy0228/ConvMemory-v2-Evidence-Reranker} (a seed-7 representative
checkpoint). The source is on GitHub at \texttt{github.com/pth2002/ConvMemory} at
tag \texttt{v0.5.0} (commit \texttt{48b80b4}). A minimal usage example:

\begin{verbatim}
from convmemory import ConvMemory
m = ConvMemory.from_pretrained("Purdy0228/ConvMemory-LoCoMo-MPNet")
m.load_evidence_reranker("Purdy0228/ConvMemory-v2-Evidence-Reranker")
ranked = m.retrieve(query=q, memories=ms, evidence_reranker="v2", top_k=10)
\end{verbatim}

The exact internal reproduction of the v363 5-seed method-level headline numbers
uses \path{experiments/v361_top10_evidence_reranker.py}. The public,
general-purpose training recipe (a user-facing entry point, not the LoCoMo-locked
harness) is \path{examples/train_evidence_reranker.py}.

\section{Discussion and Future Work}
\label{sec:discussion}

The three future-work directions named in the v1 report --- multi-backbone
checkpoint distribution, an end-to-end agent benchmark integrating ConvMemory
into a full pipeline, and broader (multi-task, multi-seed) CCGE-LA training ---
all remain open. v2 addresses a separate, unanticipated direction. Natural next
steps specific to v2 include: validating the
recall-preserving cascade on non-LoCoMo conversational-memory data; measuring
whether v2's retrieval-stage MRR/H@1 gains translate into downstream answer
quality in an end-to-end agent; and studying how far the protected-prefix width
(here, ten) can be widened before the cost advantage over a full-pool
cross-encoder erodes.

\appendix

\section{Full Latency Protocol}
\label{app:latency}

This appendix documents the latency measurement behind Table~\ref{tab:cost} in
\S\ref{sec:cost}. As in the v1 report, the intent is not benchmark-grade latency
publication but a reproducible cost-frontier comparison and an explicit statement
of what each figure includes and excludes.

\paragraph{Setup.} Measurements were taken on a single NVIDIA GeForce RTX 4080
SUPER. The probe times 200 queries after a warmup pass, on \texttt{cuda}. The v1
candidate pool is the dense MPNet top-500; v2 reranks only the protected v1 top-10.

\paragraph{What each row measures.}
\begin{itemize}
    \item \textbf{v1 top500} (16.779 ms/q): the v1 reranking path over the
    top-500 pool, producing the top-10 set v2 will reorder.
    \item \textbf{v2 (top10 only)} (11.867 ms/q): ten cross-encoder pair scorings over the protected top-10, excluding the upstream v1 cost.
    \item \textbf{v1 + v2} (28.646 ms/q): the deployed opt-in path; roughly the
    sum of the previous two, i.e.\ $1.71\times$ v1 alone.
    \item \textbf{mxbai top10} (79.770 ms/q): mxbai-large restricted to the same
    protected ten candidates, the set-matched cross-encoder control.
    \item \textbf{mxbai top500} (1960.248 ms/q): mxbai-large over the full top-500
    pool, the high-cost full-pool reference.
\end{itemize}

\paragraph{Ratios.} mxbai\_top500 / (v1+v2) $= 68.43\times$; mxbai\_top500 / v1
$= 116.83\times$; mxbai\_top10 / v2\_top10 $= 6.72\times$; (v1+v2) / v1
$= 1.71\times$.

\paragraph{Caveats.} All figures are from a single run on a single consumer GPU
(RTX 4080 SUPER) with 200 timed queries after warmup; data-center GPUs, smaller
GPUs, CPU inference, or kernel-fused inference servers would shift these numbers
and should be measured on the target hardware. The cross-encoder baselines are
off-the-shelf checkpoints at default sequence length. These are indicative
cost-frontier measurements, not benchmark-grade latency claims.

\section{Released Checkpoint Provenance}
\label{app:provenance}

The relationship between the published checkpoint and the headline numbers
warrants a precise statement, because the two are easy to conflate.

\begin{itemize}
    \item The original v361 5-seed run did not save its per-seed checkpoints.
    There is therefore no archived single checkpoint that ``is'' the 5-seed
    headline.
    \item The checkpoint published on the Hugging Face Hub
    (\path{Purdy0228/ConvMemory-v2-Evidence-Reranker}) is a \textbf{seed-7
    representative} checkpoint, exported after v0.5.0 packaging using the same
    v361 gold-only recipe.
    \item Its manifest records the recipe: teacher weight $0.0$ (gold-only),
    \texttt{top\_k} $=10$, \texttt{max\_length} $=256$, and cross-encoder backbone
    \path{cross-encoder/ms-marco-MiniLM-L-6-v2}, with source experiment
    \path{v361_top10_evidence_reranker.py} and training target
    ``gold-only listwise retrieval cross-entropy.''
    \item Consequently, the released checkpoint \textbf{implements the method};
    the headline metrics (FULL MRR 0.6560, etc.) are \textbf{method-level 5-seed
    estimates}, not the measured score of this one checkpoint.
\end{itemize}

The 0.6560 figure is therefore a method-level estimate, not the individual
score of the released checkpoint. A user evaluating the single downloaded
checkpoint should expect a single-seed result consistent with, but not identical
to, the 5-seed method-level estimate.

\section{Source-of-Truth Checklist}
\label{app:sources}

Every quantitative claim in this report traces to one of the files in
Table~\ref{tab:sources}. Where exact bibliographic metadata for external model
cards could not be verified, those models are cited as footnoted URLs rather than
as fabricated references.

\begin{table}[htbp]
\centering
\caption{Mapping from claim type to its source-of-truth file. The public package
and test files (\texttt{convmemory/evidence\_reranker.py},
\texttt{tests/test\_evidence\_reranker.py}) are in the GitHub repository at tag
\texttt{v0.5.0}; the \texttt{results/} verifier packets and the internal
experiment script are local source-of-truth artifacts retained by the author and
available on request, not part of the public tag.}
\label{tab:sources}
\renewcommand{\arraystretch}{1.2}
\begin{tabular}{>{\raggedright\arraybackslash}p{0.34\textwidth} >{\raggedright\arraybackslash}p{0.58\textwidth}}
\toprule
Claim type & Source file \\
\midrule
v1/v2 headline metrics & \path{results/v363_v2_verifier_packet/headline_table.csv} \\
bootstrap CIs (v2$-$v1) & \path{results/v363_v2_verifier_packet/bootstrap_v2_minus_v1.csv} \\
v361 method details & \path{experiments/v361_top10_evidence_reranker.py} \\
v1 anchor reproduction & \path{results/v359_v1paper_compatible_v2/REPORT.md} \\
recall-preserving merge & \path{results/v360_recall_preserving_v2_merge/REPORT.md} \\
latency / cost & \path{results/v362_v361_latency_cost/REPORT.md} \\
load-bearing ablation & \path{results/v364_v361_token_evidence_ablation/REPORT.md} \\
checkpoint provenance & \path{results/v365_v05_evidence_reranker_checkpoint/seed_7/MANIFEST.json} \\
public API guard & \path{convmemory/evidence_reranker.py} \\
isolation tests & \path{tests/test_evidence_reranker.py} \\
\bottomrule
\end{tabular}
\end{table}

\end{document}